\documentclass[11pt]{article}

\usepackage[final]{acl}

\usepackage{times}
\usepackage{latexsym}

\usepackage[T1]{fontenc}

\usepackage[utf8]{inputenc}

\usepackage{microtype}

\usepackage{inconsolata}

\usepackage{graphicx}
\usepackage{amsmath}
\usepackage{amsfonts}  
\usepackage{amssymb}   
\usepackage{booktabs}
\usepackage{tabularx} 
\usepackage{graphicx} 

\usepackage{multirow} 
\usepackage[table]{xcolor}
\newcommand{\method}{OCR-Memory}

\title{OCR-Memory: Optical Context Retrieval for Long-Horizon Agent Memory}

\author{
Jinze Li\textsuperscript{1},
Yang Zhang\textsuperscript{2,$\dagger$},
Xin Yang\textsuperscript{3},
Jiayi Qu\textsuperscript{4},
Jinfeng Xu\textsuperscript{1}, \\
\textbf{Shuo Yang\textsuperscript{1},
Junhua Ding\textsuperscript{2},
Edith Cheuk-Han Ngai\textsuperscript{1,$\dagger$}}
\\
\textsuperscript{1}The University of Hong Kong 
\textsuperscript{2}University of North Texas \\
\textsuperscript{3}University of Tsukuba  
\textsuperscript{4}Yonsei University \\
\texttt{\{lijinze-hku, shuo.yang, jinfeng\}@connect.hku.hk} \\
\texttt{\{yang.zhang, Junhua.Ding\}@unt.edu} \\
\texttt{s2330128@u.tsukuba.ac.jp}, \texttt{jiayiqu12@gmail.com} \\
\texttt{chngai@eee.hku.hk}
}

\begin{document}
\maketitle
\begin{abstract}

Autonomous LLM agents increasingly operate in long-horizon, interactive settings where success depends on reusing experience accumulated over extended histories. However, existing agent memory systems are fundamentally constrained by text-context budgets: storing or revisiting raw trajectories is prohibitively token-expensive, while summarization and text-only retrieval trade token savings for information loss and fragmented evidence.
To address this limitation, we propose Optical Context Retrieval Memory (\textbf{\method{}}), a memory framework that leverages the visual modality as a high-density representation of agent experience, enabling retention of arbitrarily long histories with minimal prompt overhead at retrieval time.
Specifically, \method{} renders historical trajectories into images annotated with unique visual identifiers.
\method{} retrieves stored experience via a \emph{locate-and-transcribe} paradigm that selects relevant regions through visual anchors and retrieves the corresponding verbatim text, avoiding free-form generation and reducing hallucination. Experiments on long-horizon agent benchmarks show consistent gains under strict context limits, demonstrating that optical encoding increases effective memory capacity while preserving faithful evidence recovery.

\end{abstract}

\section{Introduction}
Large language models (LLMs) are transforming AI systems from static question-answering engines into autonomous agents that interact with tools and users over extended periods~\cite{xi2023rise, wang2023voyager, park2023generative}. In many real-world deployments, e.g., web automation and mobile app operation~\cite{yang2024appagent}, an agent does not solve a single isolated problem but rather handles a continuous stream of tasks. In this setting, performance depends not only on the agent’s within-task reasoning but 
on its ability to accumulate experience across completed episodes and reuse it when similar new tasks arise.

However, effectively retaining such long-horizon memory remains a fundamental challenge due to the inherent conflict between the richness of experience and the constraints of LLMs. During extended interactions, agents generate extensive histories containing reasoning traces, tool invocations, and environmental feedback, containing details that are ideally kept completely for future reference. Yet, the finite context window of LLMs makes it impractical to store or revisit these high-fidelity trajectories in their entirety~\cite{liu2024lost}. Consequently, existing approaches are forced to compress past experience via summarization or abstraction~\cite{packer2023memgpt, zhong2024memorybank}. This compromise often results in the loss of structural, temporal, or procedural details that are critical for complex downstream tasks such as debugging, error analysis, or multi-step planning.

To address these limitations, we investigate the visual modality as a superior alternative for representing agent experience. Recent advancement~\cite{wei2025deepseek} demonstrates that dense textual content can be encoded into visual tokens that consume substantially less context than raw text, while crucially maintaining the full fidelity of the original information. This property suggests that visual representations can serve as a high-density, loss-free medium for long-term memory.

In light of this observation, we propose Optical Context Retrieval Memory (\textbf{\method{}}), a framework that stores an agent’s complete interaction trajectories as images. By encoding extensive interaction history into a small number of visual tokens, \method{} 
avoids the trade-off between memory capacity and information completeness, enabling the scalable storage of arbitrarily long histories without lossy summarization or truncation.

To retrieve precise information from this visual store, \method{} employs a \emph{locate-and-transcribe} mechanism. 
Specifically, interaction logs are rendered into images annotated with unique visual anchors such as indexed bounding boxes. 
When the agent requires historical context, the optical retrieval module scans these visual representations to predict the specific indexes of relevant segments, rather than generating a free-form textual response. 
Once identified, the corresponding original text is deterministically fetched from the database based on the selected indexes. This design  decouples context understanding from evidence generation, allowing the agent to efficiently retrieve from massive visual histories with minimal token cost while ensuring that the retrieved context remains verbatim and hallucination-free.
To mimic the vivid-to-fuzzy property of human memory, we introduce an age-aware adaptive-resolution scheme that progressively stores older trajectory images as low-resolution thumbnails, keeping the visual-token cost of long interaction histories manageable. 
Crucially, these low-resolution thumbnails preserve the semantic gist, sufficient for retrieval despite the loss of fine details. When such a \emph{fading} memory is identified as relevant, our active-recall up-sampling serves as a memory refresh mechanism: it restores the image to high fidelity, ensuring useful context is available in full detail for subsequent interaction steps.

Empirical evaluations on the Mind2Web~\cite{deng2023mind2web} and AppWorld~\cite{chi2024appworld} benchmarks demonstrate that \method{} consistently outperforms strong baselines, establishing a new state-of-the-art for long-horizon agent tasks. Furthermore, extensive ablation studies validate the effectiveness of our design, confirming that the visual anchoring mechanism significantly reduces hallucination risks while maintaining robustness under strict token budgets.

In summary, the key contributions of this study are summarized as follows:
\begin{itemize}
    \item We propose OCR-Memory, the first memory framework to store an agent’s interaction history in the image modality, enabling complete retention of episodic experience under a limited context window.
    \item We introduce a \emph{Locate-and-Transcribe} retrieval pipeline to mitigate hallucinations in optical retrieval. By using indexed visual anchors, retrieval becomes explicit index selection rather than free-form generation, and the original text is then deterministically recovered from external logs using the selected indices. 
    \item We design adaptive resolution and active-recall up-sampling to look far with manageable token cost, while preserving high fidelity for salient memories.
    \item We conduct comprehensive experiments to demonstrate \method{}’s superior performance compared to existing methods.
\end{itemize}

\begin{figure*}[htbp]
    \centering
    \includegraphics[width=0.9\textwidth]{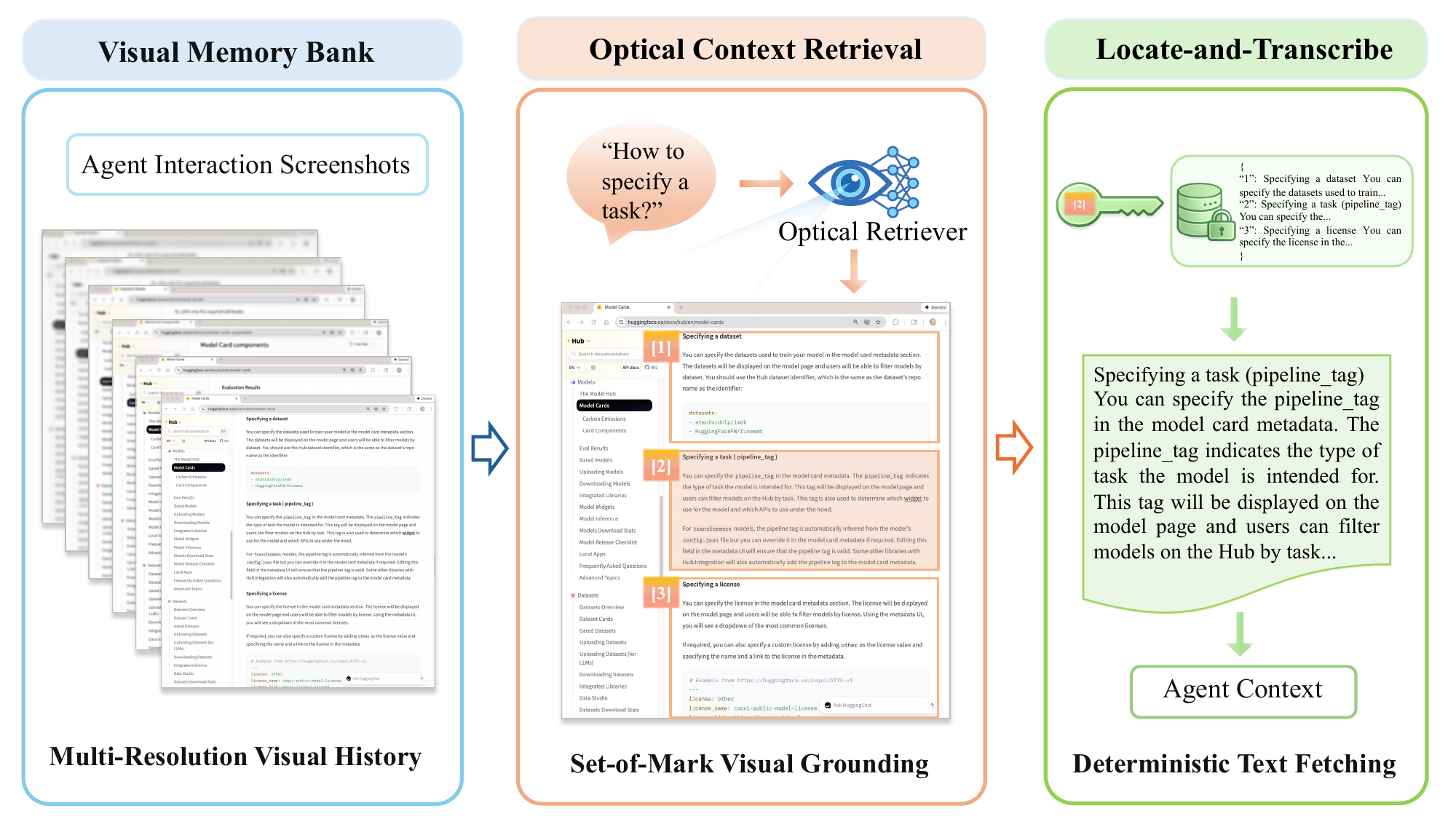} 
    \caption{Overview of the \method{}. The system enables long-horizon agent memory by storing interaction histories as compressed multi-resolution images (left). To retrieve information, we employ a Locate-and-Transcribe paradigm: the model scans the visual history annotated with Set-of-Mark (SoM) visual anchors (center) to predict the index of relevant segments. Finally, the verbatim text corresponding to the selected index is deterministically fetched (right), avoiding generation-based hallucinations and minimizing token usage.}
    \label{fig:overview}
\end{figure*}

\section{Related Work}
Recent work on agent memory falls into three paradigms: retrieval-based memory, experience abstraction, and context compression. A shared challenge is supporting long-horizon tasks under finite context constraints~\cite{xi2023rise, zhou2024navigating, liu2024lost, bai2023longbench}.

\textbf{Retrieval-Based Memory Systems.} Existing retrieval-based approaches store past interactions externally and fetch relevant fragments at inference time~\cite{lewis2020retrieval, park2023generative, packer2023memgpt, hu2024chatdb, sarthi2024raptor, zhong2024memorybank, shinn2023reflexion, asai2023selfrag, yan2025memory}. This design effectively expands usable context via semantic retrieval and lightweight memory management, but its core dependency on similarity matching can be brittle: retrieved snippets may be topically related yet logically irrelevant, especially when tasks hinge on causality or long-range dependencies~\cite{yan2025memory}.

\textbf{Experience Abstraction.} Another line of work compresses trajectories into reusable skills, workflows, or procedural knowledge to reduce future reasoning cost~\cite{wang2023voyager, zhu2023ghost, wang2024agent, wen2023dilu, zhao2023expel, hong2024metagpt, sumers2024coala}. While abstraction improves efficiency by replacing verbose logs with higher-level rules, it can discard crucial low-level details (e.g., exact error messages, intermediate states, or nuanced dialogue turns), which are often necessary for debugging, faithful retrospection, and grounded decision-making.

\textbf{Context Compression.} Instead of selecting or abstracting history, recent methods aim to compress the context itself via latent memory representations, learned compression policies, token pruning, or streaming-friendly inference mechanisms~\cite{zhang2025memgen, kang2025acon, jiang2023llmlingua, li2023selective, xiao2023streamingllm}. However, text-centric compression inevitably trades off compression ratio against information fidelity; the risk is amplified in multimodal settings where visual layouts and structural cues are essential but easy to lose under pure textual summarization~\cite{yang2024appagent, hong2024cogagent}.

\section{Preliminaries}
\paragraph{Agent execution and trajectories.}
We consider an LLM-based agent that solves tasks through multi-step interaction with an environment (tools, files, APIs, user messages). For an episode $e$, the agent observes an input query $q^{(e)}$ and generates a solution trajectory $\tau^{(e)}$:
\begin{equation}
\tau^{(e)}=\{x^{(e)}_1, x^{(e)}_2, \ldots, x^{(e)}_{T_e}\},
\end{equation}
where each element $x_t$ may be a user turn, an intermediate reasoning trace, a tool invocation, or a tool result. 

\paragraph{External memory usage.}
We formalize ``agent memory'' as an external store $\mathcal{M}$ that accumulates past trajectories, and a retrieval module that selects a small, task-relevant subset to inject back into the agent prompt. Concretely, after each episode we write $\tau^{(e)}$ into $\mathcal{M}$. When a new query $q$ arrives, a retrieval function $g_{\theta}$ reads $\mathcal{M}$ and returns relevant evidence $E$ that will be inserted into the agent's prompt:
\begin{equation}
\begin{aligned}
E &= g_{\theta}(q,\mathcal{M}).
\end{aligned}
\end{equation}
The agent then performs the actual reasoning and tool use conditioned on relevant evidence $E$ and query $q$.

\paragraph{Visual Encoding.}
DeepSeek-OCR~\cite{wei2025deepseek} treats \emph{contexts optical compression} as an end-to-end image encoding procedure that maps a document image into a compact sequence of latent embeddings. 
Formally, given an input image $\mathbf{I}\in\mathbb{R}^{H\times W\times 3}$, the image encoder produces compressed embeddings
\begin{equation}
\mathbf{Z}=f_{\mathrm{enc}}(\mathbf{I})\in\mathbb{R}^{n(r)\times d_{\mathrm{latent}}},
\label{eq:deepencoder}
\end{equation}
where $r$ denotes the resolution mode and $n(r)$ is the resulting compressed-token budget. 
To enable controllable compression ratios, $f_{\mathrm{enc}}$ equips with multiple resolution modes, which is preset by the chosen input size:
\begin{equation}
n(r)\in\{64,\,100,\,256,\,400\},
\label{eq:native-token-budget}
\end{equation}
corresponding to $512\!\times\!512$, $640\!\times\!640$, $1024\!\times\!1024$, and $1280\!\times\!1280$ inputs, respectively.

\section{OCR-Memory}
\label{subsec:ocr-memory}
We now introduce Optical Context Retrieval for Agent Memory (\method{}), a paradigm that shifts memory storage and retrieval from text domain to the image domain, so that a memory module with a limited text context window can still consult arbitrarily long histories with minimal token overhead.
Our main modification is to store trajectories in $\mathcal{M}$ as images rather than raw text.
This design decouples the storage and retrieval of episodic history from the reasoning process of the primary agent by introducing a specialized optical model that is used solely for context retrieval.
By identifying relevant segments via visual marks instead of generating free-form text, the model significantly reduces hallucination risk and generation latency.

Formally, we maintain an external memory bank
\begin{equation}
\begin{aligned}
\mathcal{M} &= \{m_i\}_{i=1}^{N}, \\
m_i &= \bigl(\mathbf{I}_i,\ \{s_{i,k}\}_{k=1}^{K_i},\ \pi_i\bigr),
\end{aligned}
\label{eq:memory-bank}
\end{equation}
where $\mathbf{I}_i$ is a rendered (marked) image of a stored trajectory chunk, $\{s_{i,k}\}$ are the corresponding text segments (stored verbatim in a deterministic log), and $\pi_i$ denotes metadata (timestamp, episode id, etc.).
Given a new query $q$, the \method{} $g_{\theta}$ reads $\mathcal{M}$ and returns a small set of supportive segments to be injected into the primary agent context:
\begin{equation}
\hat{\mathcal{S}}(q)=g_{\theta}(q,\mathcal{M}),
\quad
E=\mathrm{Fetch}(\hat{\mathcal{S}}(q),\mathcal{M}),
\label{eq:retrieve-fetch}
\end{equation}
where $\hat{\mathcal{S}}(q)$ is an index set of selected segments and $\mathrm{Fetch}(\cdot)$ maps indices to the exact stored segment texts.
Importantly, $g_{\theta}$ is not required to answer $q$, it is optimized only to retrieve evidence that improves the downstream agent's success rate under a finite token budget.

\paragraph{Locate-and-Transcribe.}
Visual retrieval directly from images often suffers from textual hallucinations, particularly due to low resolution or blurriness. We mitigate this via a \emph{Locate-and-Transcribe} paradigm that transforms the task from free-form generation into precise pointer selection.
Specifically, we employ Set-of-Mark (SoM) prompting, where each text segment $s_{i,k}$ in a memory image $i$ is highlighted with a red bounding box and annotated with a unique numerical ID $k\in [1,K_i]$.
In this \emph{listwise} setting, our \method{} model acts strictly as a \emph{relevance extractor}, outputting a binary relevance vector for a given image $i$ with $K_i$ segments:
\begin{equation}
\hat{\mathbf{y}}_i(q) = \big(\hat{y}_{i,1}, \ldots, \hat{y}_{i,K_i}\big) \in \{0,1\}^{K_i}
\end{equation}
We ensure strict formatting by constraining label tokens to be either ``0'' or ``1'', where $\hat{y}_{i,k}=1$
indicates that segment $k$  in image $i$  is selected.
Collecting these positive predictions across all $N$ memory images yields the global index set 
\begin{equation}
\hat{\mathcal{S}}(q) = \{(i,k) \mid \hat{y}_{i,k}=1 \},
\end{equation}
where
\begin{equation}
 1 \leq i \leq N, \quad 1 \leq k \leq K_i  .
\end{equation}
This ``index-only'' output is substantially faster and allows the system to deterministically ``transcribe'' content by fetching exact stored texts from the memory $\mathcal{M}$:
\begin{equation}
E = \mathrm{Fetch}\big(\hat{\mathcal{S}}(q),\mathcal{M}\big) = \bigoplus_{(i,k)\in \hat{\mathcal{S}}(q)} s_{i,k},
\label{eq:concat-evidence}
\end{equation}
where $\oplus$ denotes concatenation under a fixed formatting template.
This separation of concerns leverages visual grounding for search while reserving the primary agent for reasoning.

\paragraph{Inference Scoring.}
While the strict binary formatting simplifies the training objective, relying solely on greedy decoding (i.e., strictly selecting segments where $\hat{y}_{i,k}=1$) poses a risk of false negatives.
To robustly capture relevant evidence, we derive calibrated relevance scores from the underlying token logits, rather than using the discrete outputs directly.
Let $z_{i,k}(1)$ and $z_{i,k}(0)$ denote the decoder logits at the label position for tokens ``1'' and ``0''.
We define the segment relevance probability as
\begin{equation}
p_{i,k}(q)=\frac{\exp(z_{i,k}(1))}{\exp(z_{i,k}(1))+\exp(z_{i,k}(0))}.
\label{eq:prob-from-logits}
\end{equation}
Adhering to the retrieval preference ``better to retrieve more than to miss,'' we adopt a recall-oriented selection rule.
For each image $i$, we construct the candidate set using a low threshold $\tau$ with a Top-$K$ fallback:
\begin{equation}
\begin{aligned}
\hat{\mathcal{S}}_i(q)=\{(&i,k)\mid p_{i,k}(q)\ge \tau\}\ \\
&\cup\ \mathrm{TopK}\big(\{p_{i,k}(q)\}_{k=1}^{K_i}, K\big).
\end{aligned}
\label{eq:threshold-topk}
\end{equation}
Here, $\text{TopK}(\cdot)$ returns the set of indices $(i, k)$ corresponding to the $K$ highest probability scores within image $i$. The union operation ($\cup$) enforces a \textit{minimum-guarantee policy}: it prioritizes high-confidence segments satisfying $p_{i,k}(q) \geq \tau$, while the Top-$K$ component ensures that at least $K$ segments are retrieved per image even when the model is uncertain (i.e., when no segments exceed $\tau$). This hybrid selection strategy robustly maintains coverage without manual tuning of instance-specific thresholds.
Finally, we obtain the global retrieved set by aggregating predictions across all memory images: $\hat{\mathcal{S}}(q)=\bigcup_{i=1}^{N}\hat{\mathcal{S}}_i(q)$.

\paragraph{Multi-Resolution Trajectories.}
To support visual retrieval, we render trajectory chunks into high-fidelity marked images, denoted as $\mathbf{I}_i^{\mathrm{hi}}$.
This rendering preserves spatial layouts that carry semantic meaning but are costly to represent in text tokens.
To emulate the ``vivid-to-fuzzy'' property of human memory, we define the memory age $\Delta t_i$ as the time elapsed since storage.
We apply a dynamic resolution policy where an older memory is assigned to a higher aging tier $\ell_i$, which in turn dictates a lower image resolution:
\begin{equation}
\ell_i=\rho(\Delta t_i),
\qquad
\mathbf{I}_i=\phi_{\ell_i}(\mathbf{I}_i^{\mathrm{hi}}),
\label{eq:dynamic-resolution}
\end{equation}
where $\rho(\cdot)$ is a monotonic mapping and $\phi_{\ell}(\cdot)$ performs downsampling based on tier $\ell$.
This ``optical forgetting'' significantly reduces the visual-token cost of long-term history.
Crucially, this degradation is reversible via \emph{Active Recall Upscaling}.
If the retrieval module identifies a relevant segment in a compressed memory item, specifically satisfying:
\begin{equation}
\exists (i,k) \in \hat{\mathcal{S}}(q) \quad \text{s.t.} \quad \ell_i > \ell_{\min},
\label{eq:upscale-condition}
\end{equation}
we instantly restore the image to its original fidelity:
\begin{equation}
\mathbf{I}_i \leftarrow \mathbf{I}_i^{\mathrm{hi}}.
\label{eq:active-recall-upscale}
\end{equation}
We do not store redundant high- and low-resolution copies for each memory item. Instead, we persist the raw text logs together with a single current image representation. When a low-resolution memory is retrieved, we re-render its high-resolution version on demand from the original logs and keep it in the active visual cache for the remainder of the episode. This design avoids duplicate storage while preserving the ability to recover full-fidelity evidence when needed.
This design mimics the natural decay of human memory: while specific characters in older, low-resolution images may be blurred, the semantic gist and general context remain discernible. When the model successfully retrieves a segment based on this ``fuzzy'' understanding, it implies the memory is currently critical.
Consequently, it acts as an adaptive filter: rarely accessed or low-utilization memories remain in a compressed low-resolution state, while frequently retrieved and highly useful information is maintained at high fidelity.

\section{Training Strategy}
\label{subsec:training-strategy}

The backbone of our method is DeepSeek-OCR.
However, the pre-trained model is optimized primarily for literal transcription and is weak at instruction-following for relevance matching.
In our setting, the model must not only \emph{read} but also \emph{judge} which passages support a query.
We therefore fine-tune the model for discriminative retrieval using a repurposed HotpotQA dataset~\cite{yang2018hotpotqa}.

\paragraph{Repurposing HotpotQA.}
A HotpotQA instance consists of a question $q$, a context of $K$ paragraphs $\{p_1,\ldots,p_K\}$ (typically $K=10$), and a set of annotated supporting facts.
We discard the textual answer and strictly supervise the model using these supporting facts.
Let $\mathcal{F}_{\text{supp}}$ denote the set of indices corresponding to the ground-truth supporting paragraphs.
We define the binary target label $y_k$ for the $k$-th paragraph as:
\begin{equation}
y_k = \mathbb{I}[k \in \mathcal{F}_{\text{supp}}],
\label{eq:hotpot-label}
\end{equation}
where $\mathbb{I}[\cdot]$ is the indicator function, which equals 1 if the condition holds and 0 otherwise.
Consequently, the target vector is 
\begin{equation}
\mathbf{y}=(y_1, \ldots, y_K) \in \{0,1\}^K
\end{equation}

We render the $K$ paragraphs into marked images with SoM identifiers:
\begin{equation}
\mathbf{I}=\mathrm{Render}\Big(\{(k,p_k)\}_{k=1}^{K}\Big).
\label{eq:hotpot-render}
\end{equation}
The training objective is to produce this correct binary vector $\mathbf{y}$.
This transforms the problem from next-token prediction for generation into next-token prediction for precise evidence retrieval.

\begin{table*}[t]
\centering
\resizebox{0.85\textwidth}{!}{%
\begin{tabular}{l|cccc|cccc}
\toprule
\multirow{2}{*}{Method} & \multicolumn{4}{c|}{Mind2Web} & \multicolumn{4}{c}{AppWorld (SR \%)} \\
 & Ele Acc  & F1 Score  & Step SR  & Task SR  & Easy  & Med  & Hard  & Avg  \\ \midrule
 Zero-Shot  & 40.1 & 46.2 & 37.9 & 2.2 & 68.7 & 36.2 & 20.9 & 41.9 \\
 \midrule
 Retrieval & 41.3 & 48.2 & 38.9 & 2.7 & 72.5 & 44.8 & 21.4 & 46.2 \\
 MemoryBank~\cite{zhong2024memorybank} & 43.8 & 49.5 & 39.2 & 3.3 & 81.3 & 50.1 & 24.9 & 52.1 \\
 AWM~\cite{wang2024agent} & 49.1 & 55.7 & 42.6 & 4.3 & 84.1 & 53.6 & 27.2 & 55.0 \\
 ACON~\cite{kang2025acon} & 48.2 & 54.1 & 41.4 & 4.1 & 84.8 & 55.1 & 28.7 & 56.2 \\ 
 \rowcolor{gray!15}
 \textbf{\method{}} & \textbf{53.8} & \textbf{59.2} & \textbf{46.1} & \textbf{4.8} & \textbf{86.2} & \textbf{57.4} & \textbf{30.8} & \textbf{58.1} \\ \bottomrule
\end{tabular}%
}
\caption{Main Results on Long-Horizon Agent Tasks. We report fine-grained metrics for Mind2Web, including Element Accuracy, Action F1 Score, Step Success Rate, and Task Success Rate. For AppWorld, we report Success Rates across three difficulty levels and the overall Average. \method{} consistently outperforms baselines, achieving significant gains in metrics requiring precise structural grounding.}
\label{tab:main_results_final}
\end{table*}

\paragraph{Optimization Objective.}
Let the constructed dataset containing input images, queries, and ground-truth binary labels be denoted as:
\begin{equation}
\mathcal{D}=\Big\{(\mathbf{I}^{(n)},q^{(n)},\mathbf{y}^{(n)})\Big\}_{n=1}^{|\mathcal{D}|}.
\label{eq:dataset-def}
\end{equation}
Since the retrieval task is inherently imbalanced where positive segments are sparse compared to negatives, we employ a weighted binary cross-entropy objective to supervise the model under teacher forcing.
Using the calibrated probability $p_{k}^{(n)}$ derived from logits (as defined in Eq.~\eqref{eq:prob-from-logits}), the loss function is formulated as:
\begin{equation}
\begin{aligned}
\mathcal{L}_{\mathrm{BCE}}(\theta) & = -  \sum_{n=1}^{|\mathcal{D}|} \sum_{k=1}^{K} \Big[  w_{+} \cdot y_{k}^{(n)} \log p_{k}^{(n)} \\
& + w_{-} \cdot (1-y_{k}^{(n)}) \log (1-p_{k}^{(n)}) \Big],
\end{aligned}
\label{eq:weighted-bce}
\end{equation}
where $y_{k}^{(n)}$ is the ground truth for the $k$-th segment in sample $n$.
To bias the model toward higher recall, we strictly set the weights such that:
\begin{equation}
w_{+} > w_{-},
\label{eq:weight-condition}
\end{equation}
which penalizes false negatives (missed evidence) more heavily than false positives.

\paragraph{Training Strategy.}
To preserve the robust visual representations learned during pre-training while adapting the model for fine-grained grounding, we adopt a partial freezing strategy.
We partition the model parameters into the vision encoder $\theta_{\mathrm{vis}}$ and the language decoder $\theta_{\mathrm{dec}}$:
\begin{equation}
\theta = \theta_{\mathrm{vis}} \cup \theta_{\mathrm{dec}}.
\label{eq:param-split}
\end{equation}
In our default setting, we freeze $\theta_{\mathrm{vis}}$ to maintain stability and update only the decoder parameters via LoRA:
\begin{equation}
\theta_{\mathrm{vis}} \leftarrow \text{fixed},
\
\theta_{\mathrm{dec}} \leftarrow \theta_{\mathrm{dec}} - \eta \nabla_{\theta_{\mathrm{dec}}}\mathcal{L}_{\mathrm{BCE}}.
\label{eq:freeze-vis}
\end{equation}
This ensures the model learns the semantic mapping from textual queries to visually grounded segment anchors (SoM marks) without destabilizing the fundamental optical recognition capabilities.

\paragraph{Resolution Curriculum for Long-Term Memory.}
During inference, the model must retrieve information from memory items that may have heavily degraded resolutions due to the aging mechanism.
However, standard training data (e.g., HotpotQA renders) typically consists of high-fidelity images.
To bridge this domain gap and simulate the multi-resolution conditions encountered during deployment, we apply a resolution curriculum during training.
For each training instance, we randomly sample a resolution tier $\ell$ and downsample the rendered image accordingly:
\begin{equation}
\ell \sim \mathrm{Categorical}(\pi),
\qquad
\tilde{\mathbf{I}} = \phi_{\ell}(\mathbf{I}),
\label{eq:resolution-aug}
\end{equation}
where $\mathrm{Categorical}(\pi)$ denotes a discrete probability distribution over the set of resolution tiers $\{1, \ldots, L\}$, with the vector $\pi$ controlling the sampling likelihood of each tier.
This augmentation strategy forces the retriever to rely on coarse-grained visual cues when fine details are unavailable, ensuring robustness under the optical compression used in our long-term memory bank.

Fine-tuning on marked images instills a precise ``look-and-select'' mechanism in the model.
Instead of generating free-form text, the model learns to strictly align the query with visually grounded blocks (SoM anchors) and output distinct indices.
By combining this discriminative capability with our recall-oriented selection rule and deterministic text fetching, the \method{} module functions as a low-latency, hallucination-free retrieval engine.
It effectively supplies high-relevance context to the downstream reasoning agent, maximizing information density without exceeding the strict token budget of the context window.

\section{Experiments}
In this section, we conduct an empirical evaluation of \method{}. Our experimental design aims to move beyond performance metrics to investigate the fundamental behavioral characteristics of optical memory mechanisms in long-horizon interaction environments.

\subsection{Experimental Setup}

\paragraph{Dataset.} 
We align with prior work~\cite{kang2025acon, wang2024agent} using Mind2Web (Cross-Task split) for web navigation and AppWorld for API interactions. For Mind2Web, we report standard success and accuracy metrics. For AppWorld, we emphasize the "Hard" subset to strictly evaluate the agent's ability to handle extensive history backtracking.

\paragraph{Baseline.} We compare \method{} against five representative baselines that span the spectrum of current memory paradigms. We establish the performance lower bound using a {Zero-Shot} setting, where the agent operates without access to historical interaction logs. To evaluate standard text-based retrieval, we include a {Retrieval} baseline that utilizes dense vector similarity to fetch historical text chunks. 
We also include three existing approaches: MemoryBank~\cite{zhong2024memorybank}, Agent Workflow Memory (AWM)~\cite{wang2024agent} and ACON~\cite{kang2025acon}. Unless otherwise specified, the context window for the memory module is strictly set to 4096 tokens by default.

\paragraph{Implementation details.} The core of our framework is built upon the DeepSeek-OCR (3B) architecture. We freeze the pre-trained image encoder to preserve its optical recognition capabilities and fine-tune only the language decoder to adapt to our specific retrieval instructions. The primary reasoning agent is instantiated using GPT-4 with a temperature of 0 to ensure reproducibility. 
We employ a dynamic multi-resolution strategy: the five most recent interaction steps are stored at a high resolution (1024 $\times$ 1024) to maintain immediate clarity, while all prior history is down-sampled to 512 $\times$ 512, with an active up-scaling mechanism triggered upon retrieval hits. 
For the Set-of-Mark prompting, we standardize visual anchors using red bounding boxes with 36pt indices to maximize the attention guidance of the vision encoder.
The language decoder is fine-tuned for 3 epochs on the HotpotQA dataset, specifically utilizing the training split of the distractor subset. 
For training, we use a weighted BCE objective with $w^{+}>w^{-}$ to emphasize recall; we set $(w^{+},w^{-})=(2.0,1.0)$. We employ a cosine learning rate schedule, setting the peak learning rate to $1e-5$ with a warm-up phase covering  10\% of the total training steps. The global batch size is maintained at 128. For the resolution curriculum, we use $L=2$ tiers and sample $\ell \sim \mathrm{Categorical}({\pi})$ with ${\pi}=[0.3,0.7]$ over $\{1024\!\times\!1024,\;512\!\times\!512\}$. These correspond to DeepSeek-OCR compressed-token budgets $n(r)\in\{256,64\}$, respectively.
For dynamic memory aging we implement $\ell_i=\rho(\Delta t_i)$ as a two-level policy: the most recent $5$ interaction steps use the high-resolution tier and all earlier history uses the low-resolution tier. More details can be seen in Appendix \ref{app:implementation_details}.

\begin{table}[t]
\centering
\resizebox{\columnwidth}{!}{%
\begin{tabular}{lccc}
\toprule
Method & Ele Acc (\%) & Step SR (\%) & Latency (s) \\ 
\midrule
{\method{} (Full)} & {53.8} & {46.1} & {1.7} \\
w/o SoM (Text Gen) & 46.5 & 39.2 & 5.3 \\
w/o SoM (BBox) & 49.2 & 44.5 & 2.1 \\
\bottomrule
\end{tabular}%
}
\caption{Ablation analysis of the Set-of-Mark (SoM) mechanism on Mind2Web. We report Element Accuracy (Ele Acc), Step Success Rate (Step SR), and average Inference Latency per retrieval step.}
\label{tab:ablation_som}
\end{table}

\subsection{Main Results}

Table \ref{tab:main_results_final} summarizes the performance of \method{} against a range of text-based memory baselines on the Mind2Web and AppWorld benchmarks under the same context-budget setting.
On Mind2Web, our method yields consistent gains across all metrics, outperforming the strong abstraction-based baseline AWM by clear margins. In particular, \method{} improves Element Accuracy from 49.1\% to 53.8\% and increases Step Success Rate to 46.1\%, achieving a state-of-the-art Task Success Rate of 4.8\%. These improvements stem from our ability to retain and recover fine-grained, long-horizon textual and structural details by encoding them into high-density visual representations that can be read back with a small number of tokens.
On AppWorld, \method{} attains the highest Average Success Rate of 58.1\%. The advantage is most pronounced on “Hard” tasks, where our method reaches 30.8\%, substantially surpassing both the standard Retrieval baseline (21.4\%) and AWM (27.2\%).
Overall, these results show that using the visual modality primarily as a compact carrier for lengthy textual histories enables precise evidence recovery with markedly reduced token consumption.

\subsection{Ablation Studies}
\label{sec:ablation}

To validate the individual contributions of our proposed components, we conduct a series of ablation experiments. 

\paragraph{Set-of-Mark Prompting.} 
We compare our full model with two variants: ``w/o SoM (Text Gen)'', which generates the relevant original text, and ``w/o SoM (BBox)'', which predicts bounding boxes.
As shown in Table~\ref{tab:ablation_som}, removing SoM leads to a clear performance drop. The text generation variant suffers from a higher hallucination rate and incurring nearly triple the inference latency due to long-sequence decoding, while bounding boxes are fast but insufficiently precise. Overall, SoM provides the optimal balance between grounding accuracy and computational efficiency.

\paragraph{Multi-Resolution Active Recall.}
We evaluate our dynamic Multi-Resolution Active Recall strategy. The novel design of \method{} is that historical trajectories can be compressed into low-resolution ``thumbnails'' and selectively upscaled once relevant. We compare this dynamic approach against two static baselines: ``Static Low-Res,'' where all history is permanently downsampled to $512 \times 512$, and ``Static High-Res,'' where all history is maintained at $1024 \times 1024$. 
As shown in Table~\ref{tab:ablation_res}, the Static Low-Res model yields the lowest token consumption but suffers a significant 6.4\% drop in Step SR, as the model fails to understand the meaning of some words. Conversely, the Static High-Res model achieves performance comparable to our method (46.5\%) but at a prohibitive token cost. Our dynamic strategy matches high-res fidelity with near low-res efficiency.
\begin{table}[h]
\centering
\resizebox{\columnwidth}{!}{%
\begin{tabular}{lccc}
\toprule
Resolution Strategy & Step SR (\%) & Task SR (\%) & Avg Tokens \\ \midrule
Static Low-Res ($512^2$) & 39.7 & 2.9 & {65} \\
Static High-Res ($1024^2$) & 46.5 & 4.9 & 256 \\
{Dynamic (Ours)} & {46.1} & {4.8} & 82 \\ \bottomrule
\end{tabular}%
}
\caption{Impact of the Multi-Resolution Active Recall strategy on Mind2Web. ``Avg Tokens'' denotes average number of visual tokens consumed per history frame.}
\label{tab:ablation_res}
\end{table}

\begin{figure}[t]
    \centering
    \includegraphics[width=\linewidth]{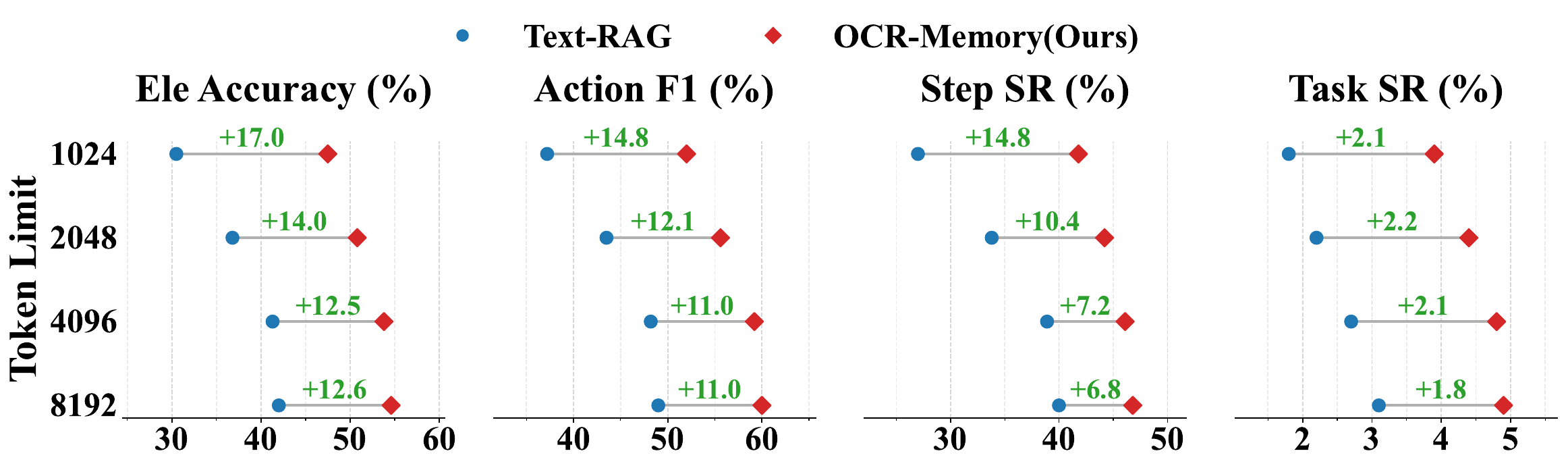}
    \caption{Performance comparison under varying context token limits.}
    \label{fig:token_limit}
\end{figure}

\paragraph{Token Constraints.} We further investigate the robustness of our framework under strict token limitations. We evaluate \method{} against a text RAG baseline across context budgets ranging from 1024 to 8192 tokens. 
As shown in Figure \ref{fig:token_limit}, \method{} consistently outperforms Retrieval method across all four Mind2Web metrics, and the performance gap widens as the token budget becomes more restrictive. Notably, even at the extreme 1024-token limit, \method{} remains functional and achieves strong Element Accuracy and Step Success Rate, whereas Text-RAG degrades substantially due to information loss. 
These results indicate that our high-density visual encoding enables token-efficient access to long-horizon history, preserving fine-grained evidence needed for precise grounding when the context budget is the primary bottleneck.

\paragraph{Retrieval Accuracy on Long-Context.}
To assess scalability, we measure the raw fidelity of our optical retrieval using the Needle-in-a-Haystack (NIAH) test from RULER benchmark. We adapted the benchmark for agents by rendering the documents into images and measuring the Recall@1 across increasing context lengths. 
As presented in Table \ref{tab:ablation_retrieval}, our method remains robust as context length increases, maintaining a retrieval accuracy of 98.5\% at 4k length and sustaining 94.1\% even when the context extends to 32k tokens. 
This accuracy comes with high efficiency: the \textit{Compression Ratio} indicates our optical encoding consistently achieves over 10$\times$ compression, effectively converting ultra-long text contexts into compact visual sequences without sacrificing semantic precision.

\begin{table}[t]
\centering
\resizebox{0.9\columnwidth}{!}{%
\begin{tabular}{c|cc}
\toprule
Context Length & Compression Ratio  & Accuracy  \\ 
\midrule
4k & 10.3$\times$ & 98.5 \\
8k & 10.2$\times$ & 97.2 \\
16k & 10.7$\times$ & 95.8 \\
32k & 10.6$\times$ & 94.1 \\ \bottomrule
\end{tabular}%
}
\caption{Results on NIAH benchmark. We report \textit{Compression Ratio} of visual tokens relative to raw text and Retrieval Accuracy (Recall@1). }
\label{tab:ablation_retrieval}
\end{table}

\begin{table*}[ht]
\centering
\small
\begin{tabular}{llccc}
\toprule
Method & Backbone & Ele Acc. (\%) & Step SR (\%) & Task SR (\%) \\
\midrule
Text Retrieval (RAG) & GPT-4 & 41.3 & 38.9 & 2.7 \\
OCR-Memory & GPT-4 & 53.8 & 46.1 & 4.8 \\
Text Retrieval (RAG) & Qwen3-32B & 35.2 & 31.5 & 1.8 \\
OCR-Memory & Qwen3-32B & 48.6 & 42.3 & 3.9 \\
\bottomrule
\end{tabular}
\caption{Backbone generalization on Mind2Web. OCR-Memory consistently outperforms text-based retrieval under both proprietary and open-source reasoning backbones, indicating that the gains arise from the memory mechanism rather than from a specific backbone.}
\label{tab:backbone_generalization}
\end{table*}

\paragraph{Backbone Generalization.} To verify that the observed improvements do not depend on GPT-4, we replace the primary reasoning agent with Qwen3-32B on Mind2Web. As shown in Table~\ref{tab:backbone_generalization}, OCR-Memory consistently outperforms text-based retrieval under both backbones. The relative gains are preserved when moving from GPT-4 to Qwen3-32B, indicating that the advantage primarily comes from the optical memory mechanism rather than from backbone-specific reasoning behavior.

\paragraph{Retrieval-Level Evaluation and Faithfulness.} Downstream task success does not directly reveal whether the retriever selects the correct historical evidence. We therefore evaluate retrieval quality on a dedicated Experience Retrieval Evaluation Subset constructed from Mind2Web. For each task, candidate memories are drawn from trajectories on the same website domain, and pseudo-gold relevant steps are annotated from the uncompressed logs. As shown in Table~\ref{tab:retrieval_eval}, OCR-Memory substantially outperforms Dense Text-RAG on Recall@1, Recall@5, Recall@10, and MRR, confirming that the learned optical retriever transfers effectively to agent-history retrieval. 

We additionally measure \textit{content-level retrieval faithfulness} at the evidence recovery stage. The free-form generative retrieval variant attains 84.3\% faithfulness, whereas OCR-Memory achieves 100.0\% because it predicts only segment indices and then deterministically fetches the associated verbatim text from the stored logs. This faithfulness metric measures whether the recovered text exactly matches stored evidence after a segment has been selected; it does not imply that every selected segment is always relevant.

\begin{table}[h]
\centering
\resizebox{\columnwidth}{!}{%
\begin{tabular}{lcccc}
\toprule
Method & Recall@1 (\%) & Recall@5 (\%) & Recall@10 (\%) & MRR \\
\midrule
Dense Text-RAG & 52.7 & 74.3 & 82.1 & 0.61 \\
OCR-Memory & 78.6 & 93.4 & 96.2 & 0.84 \\
\bottomrule
\end{tabular}
}
\caption{Retrieval-level evaluation on the Experience Retrieval Evaluation Subset from Mind2Web. OCR-Memory substantially improves retrieval relevance over Dense Text-RAG.}
\label{tab:retrieval_eval}
\end{table}

\paragraph{System Efficiency Profile.} OCR-Memory improves the utilization of the scarcest resource in long-horizon agent systems---the reasoning context window---while trading off moderate retrieval latency and higher disk usage. As shown in Table~\ref{tab:system_efficiency}, under continuous logging on Mind2Web, OCR-Memory reduces the text tokens injected into the reasoning LLM from 3,980 to 596 per step, a 6.7$\times$ reduction, at the cost of increasing disk storage per episode from 18 KB to 1.47 MB and retrieval latency from 0.3 s to 1.7 s. These results show that OCR-Memory is not universally cheaper across all resources; rather, it deliberately shifts cost away from scarce reasoning tokens toward comparatively cheaper storage and pre-processing.

\begin{table}[t]
\centering
\small
\resizebox{\columnwidth}{!}{%
\begin{tabular}{lccc}
\toprule
Method & Disk / Episode & Text Tokens / Step & Retrieval Latency / Step \\
\midrule
Text-RAG & 18 KB & 3,980 & 0.3 s \\
OCR-Memory & 1.47 MB & 596 & 1.7 s \\
\bottomrule
\end{tabular}
}
\caption{System efficiency profile on Mind2Web under continuous logging. OCR-Memory trades storage and retrieval latency for a substantial reduction in reasoning-context tokens.}
\label{tab:system_efficiency}
\end{table}

\section{Conclusion}

In this paper, we introduce \method{}, an agent memory framework that represents agent interaction trajectories as a visual stream to overcome the limitations of finite text context windows. 
\method{} performs optical context retrieval over its visual-modality memory conditioned on the input query, and returns relevant supporting evidence that helps solve the task.
By utilizing a small number of visual tokens to efficiently represent historical trajectory, our selective locate-then-transcribe mechanism precisely identifies supporting facts and recovers them verbatim, ensuring the downstream agent receives lossless and hallucination-free evidence.
Extensive evaluations demonstrate that \method{} consistently outperforms existing methods, and remains particularly robust on long-horizon tasks under tight context-window budgets.

\section*{Limitations}
Despite the effectiveness of OCR-Memory, we acknowledge several limitations. First, unlike training-free retrieval baselines, our framework requires fine-tuning a specialized optical retrieval model, which incurs additional training resource overhead. Second, the process of rendering interaction logs into images is computationally more expensive than direct text storage, and storing visual histories inevitably consumes more disk space than raw text logs. Finally, deploying the system imposes an extra memory footprint, as the parameters of the vision encoder must be maintained in memory alongside the primary language model.

\section*{Ethical Considerations}
This work does not involve human subjects, sensitive personal data, or any proprietary datasets. All datasets used in this study are publicly available and commonly used in prior research works. We have taken care to ensure that our methods and results do not raise safety, privacy, or fairness concerns.

\section*{GenAI usage disclosure.}
Generative AI tools were used for typo revising, and were not used for method design or experimental analysis.

\bibliography{custom}

\newpage
\appendix

\section{More Implementation Details}
\label{app:implementation_details}

In this section, we provide comprehensive hyperparameters and configuration details to ensure the reproducibility of our experiments.

\paragraph{Model Architecture and LoRA Configuration.}
We freeze the vision encoder ($\theta_{\text{vis}}$) of the DeepSeek-OCR (3B) and fine-tune the language decoder ($\theta_{\text{dec}}$) via LoRA. Specifically, we apply LoRA adapters to the query, key, value, and output projection layers (\texttt{q\_proj}, \texttt{k\_proj}, \texttt{v\_proj}, \texttt{o\_proj}) with a rank $r=16$, a scaling factor $\alpha=32$, and a dropout rate of 0.05. 

\paragraph{Training Hyperparameters.}
We employ the AdamW optimizer for model training, utilizing $\beta_1=0.9$, $\beta_2=0.95$, and a weight decay of 0.1. To maintain training stability, we apply gradient clipping with a max norm of 1.0. 

\paragraph{Inference and Retrieval Logic.}
The \textit{Locate-and-Transcribe} mechanism balances precision and recall through a relevance threshold $\tau$ and a fallback strategy. We set $\tau=0.4$, automatically retrieving any segment with a predicted relevance probability $p_{i,k}(q) \ge 0.4$. To ensure the agent receives minimal context even when confidence is low, we implement a Top-$K$ fallback with $K=5$ if no segments exceed $\tau$. strictly adhering to the token budget, we cap the total number of retrieved text segments at 20 per query, prioritizing segments with higher $p_{i,k}(q)$ scores when this limit is exceeded.

\paragraph{Set-of-Mark (SoM) Rendering.}
To ensure consistency between training (HotpotQA) and inference benchmarks, we adopt specific rendering specifications for visual grounding. Bounding boxes are drawn in red (RGB: 255, 0, 0) with a 3-pixel line width. Text indices use a bold 36pt sans-serif font (Arial), rendered as white text on a red background to maximize contrast for the vision encoder. Prior to feeding into the model, all images are resized to the target resolution using bicubic interpolation.

\paragraph{State Persistence in Active Recall.} Regarding the memory restoration process described in Eq.~(15), we implement the resolution update as a persistent state change. Specifically, once a low-resolution memory item is retrieved by the \textit{Locate-and-Transcribe} mechanism, it is restored to the high-resolution tier ($\mathbf{I}_i^{hi}$) and exempted from the aging decay function $\rho(\cdot)$ for the remainder of the episode. This implementation ensures that critical historical evidence, once re-activated as relevant, remains accessible in maximum fidelity for all subsequent interactions, effectively preventing valid memories from reverting to a low-fidelity state.

\end{document}